Article

# Enhancing Autonomous Vehicle Navigation with a Clothoid-Based Lateral Controller

**Aashish Shaju, Steve Southward and Mehdi Ahmadian ***

Department of Mechanical Engineering, Virginia Tech, Blacksburg, VA 24061, USA; aashish00@vt.edu (A.S.); scsouth@exchange.vt.edu (S.S.)
* Correspondence: ahmadian@vt.edu

**Featured Application: Path planning, navigation, and lateral control for autonomous vehicles. Development of control algorithm for truck platooning**.

**Abstract:** This study introduces an advanced lateral control strategy for autonomous vehicles using a clothoid-based approach integrated with an adaptive lookahead mechanism. The primary focus is on enhancing lateral stability and path-tracking accuracy through the application of Euler spirals for smooth curvature transitions, thereby reducing passenger discomfort and the risk of vehicle rollover. An innovative aspect of our work is the adaptive adjustment of lookahead distance based on real-time vehicle dynamics and road geometry, which ensures optimal path following under varying conditions. A quasi-feedback control algorithm constructs optimal clothoids at each time step, generating the appropriate steering input. A lead filter compensates for the vehicle's lateral dynamics lag, improving control responsiveness and stability. The effectiveness of the proposed controller is validated through a comprehensive co-simulation using TruckSim® and Simulink®, demonstrating significant improvements in lateral control performance across diverse driving scenarios. Future directions include scaling the controller for higher-speed applications and further optimization to minimize off-track errors, particularly for articulated vehicles.

## 1. Introduction

In the field of autonomous vehicle technology, the need for robust and precise navigation systems is crucial. Central to this pursuit is the refinement of lateral control mechanisms, which are critical in ensuring that vehicles can safely and efficiently navigate complex and dynamic environments. Lateral control, responsible for the vehicle's steering and trajectory following capabilities, is a fundamental aspect that directly influences safety, efficiency, and passenger comfort. As autonomous vehicles become increasingly prevalent, the demand for sophisticated lateral control strategies escalates. These systems must not only handle the intricate task of real-time decision making and path correction but also anticipate and adapt to unpredictable changes in road conditions and traffic dynamics. The introduction of a clothoid-based lateral controller represents a leap forward in this domain. Leveraging the mathematical properties of clothoids or Euler spirals promises to provide a control method with smoother transitions during lane changing, and hence, minimizing abrupt steering inputs and enhancing a vehicle's stability and passenger experience. This research delves into the optimization of such a controller, aiming to further bridge the existing gap in autonomous vehicle navigation through improved accuracy, reliability, and adaptability.

Before addressing path tracking, the lateral control of a ground vehicle encompasses an essential aspect known as local path planning. Local path planning involves the generation of trajectories that connect the vehicle's current coordinates to its intended destination. Within this domain, a widely adopted and effective strategy for autonomous vehicle navigation is waypoint navigation. This approach leverages GPS technology to determine the vehicle's position and subsequently generates paths between the vehicle and its destination. Strategies are then devised to follow these generated paths (path tracking) and guide the vehicle to its intended endpoint. The following paragraphs delve into the literature and contemporary techniques relevant to path generation and tracking, collectively constituting waypoint navigation for autonomous vehicles.

An early attempt to find the shortest paths between two known points was developed by Dubin in 1957 [1]. These paths, termed Dubins paths, represent the shortest curves connecting two points within a two-dimensional Euclidean plane. They adhere to curvature constraints and feature prescribed initial and terminal tangents. However, a limitation of Dubins paths lies in their constant curvature segments, which do not accurately model the variable curvature experienced by a moving vehicle. Another prevalent method for establishing trajectories between waypoints involves the use of splines [2], with cubic splines and B-splines being commonly employed. This thesis extends prior research by employing clothoids [3] for path generation, addressing issues related to steering profile smoothness and mitigating undesirable jerk in the steering wheel. This is achieved by ensuring continuity in curvature, a vital characteristic of clothoids [3,4].

Once a path has been generated, the subsequent challenge lies in guiding the vehicle along this path with minimal tracking errors and steering jerks [5]. Various control strategies, including pure pursuit [6–10], PID [11], and MPC [12,13], etc., have been proposed to address this issue. Among the earliest path-tracking strategies introduced is the pure pursuit method, which involves fitting a semicircle through the vehicle's current configuration and a point on the reference path located ahead of the vehicle at a distance referred to as the lookahead distance. Carnegie Mellon University pioneered the application of this approach to a vehicle in 1985 [14]. Subsequently, numerous studies [7–9,15] have further developed and enhanced this strategy, introducing their own variations. However, despite its advantages, the pure pursuit method faced challenges in accurately tracking the vehicle, particularly around tight corners, leading to comparatively high tracking errors when compared to other control strategies in this field. A key drawback was the tuning of the lookahead distance, as an improperly set lookahead distance could result in steering oscillations for the vehicle.

The following paragraphs focus on the different types of lateral control methods in ground-based autonomous vehicles, including PID-based control, model predictive control (MPC), sliding mode control (SMC), nonlinear controls [16,17], and AI/learning-based controls [18]. Biswas et al. [19], and Kebbati et al. [20] provide a comprehensive review of recent advancements in the lateral control of autonomous vehicles, summarizing various approaches and techniques that have been proposed in the literature. One of the key challenges in the development of CAVs (Connected Autonomous vehicles) is improving the existing lateral control algorithms to decrease cross-track error, both at low and high speeds [19]. Additionally, for articulated vehicles such as semi-tractor trailers, reducing off tracking while tracking a given trajectory is also a crucial aspect to consider.

PID-based control is a widely used method for lateral control in autonomous vehicles. Gai-Ning et al. [21] propose an adaptive PID neural network for lateral tracking control for intelligent vehicles. They established a vehicle dynamics model based on transfer functions and employed a PID controller to regulate the steering angle. Dong [22] introduced a fractional-order PID control algorithm for lateral control, demonstrating its effectiveness in overcoming tracking errors and maintaining comfort, stability, and safety. In another study [23], a deep learning (DL) model was designed to replicate the behavior of an adaptive proportional–integral–derivative (PID) controller. The DL model incorporated various factors, including communication latency, packet loss, communication range, as well as considerations pertaining to reliability, robustness, and security. In the context of vision-based autonomous vehicles, a nested PID steering control system was devised to facilitate lane keeping on roads with uncertain curvature [11]. It was designed using PID control on the lateral offset to counteract disturbances caused by curvature variations over time. Similarly, a separate investigation [24] on lateral control in articulated mining vehicles implemented a PID-based steering controller. The framework's performance was assessed for its ability to track predefined paths, even in the presence of disruptions such as uneven terrain and external forces.

Model predictive control (MPC) is another control architecture which has gained significant attention in recent years due to its ability to handle complex control problems. Yu et al. [12] provides a comprehensive review of MPC and its applications in both a single and multiple autonomous ground vehicles (AGVs). They discuss the advantages of MPC in terms of trajectory planning, collision avoidance, and control management. Devaragudi [25] presented an MPC approach for longitudinal and lateral control with real-time local path planning for obstacle avoidance. A recent study [26] introduced a hybrid control system combining (ACC) adaptive cruise control with (NMPC) nonlinear model predictive control to enhance energy efficiency for autonomous electric vehicles on curvy roads while ensuring safe vehicle spacing. Another work [13] is centered on enhancing vehicle stability during high-speed scenarios by amalgamating steady-state response and MPC techniques. Furthermore, the research of [27] delves into the comparison between decoupled and coupled adaptive MPC-based control for lateral guidance. In the coupled controller, adaptive model predictive control (MPC) is employed to manage both lateral and longitudinal control aspects. In contrast, the decoupled strategy separates the longitudinal dynamics, managed with a PID controller, from lateral dynamics, initially overseen by MPC and subsequently by a lateral controller. This innovative approach harnesses adaptive MPC to guide lateral movements and regulate speed by adjusting the front-wheel steering angle and acceleration/deceleration.

Sliding mode control (SMC) is another popular approach for lateral control in autonomous vehicles. Ro et al. [28]. introduced a model reference control scheme for a four-wheel steering system, incorporating sliding mode control theory. The Sliding Mode Four-Wheel Steering (SM4WS) method not only enhances directional stability and responsiveness but also exhibits robust disturbance rejection capabilities, particularly concerning unexpected side winds. Wang et al. [29] investigated nonlinear path-tracking systems for autonomous vehicles equipped with four-wheel steering (4WS). Their research centers on the development of a robust adaptive sliding mode controller. To account for control input nonlinearity and the parameter-varying nature of tire cornering stiffness during extreme handling scenarios, a Takagi-Sugeno (T-S) fuzzy model is employed. Zhang et al. [30] delves into research concerning path planning and path-tracking control for autonomous vehicles, utilizing an improved Artificial Potential Field (APF) approach and SMC. This controller integrates lateral and heading errors, enhancing the sliding mode function, and elevating path-tracking precision.

Recently, several research studies have investigated AI-based [18,31–33] lateral control methods for autonomous cars and trucks. Zhao et al. [31] addresses challenges in lateral motion control for Automated Highway Systems, focusing on lane keeping and changing. It introduces a multi-model fuzzy controller, with four local controllers for varying speeds, ensuring smooth transitions. Swain et al. [34] enhances lateral control for Automated Vehicles (AVs) by combining sliding mode control and radial basis function neural networks, and it effectively mitigates chattering and reduces the impact of external disturbances in diverse conditions. Another work [35] explores AI-based lateral control for automated vehicles, developing two controllers—one using fuzzy logic and another using ANFIS with expert driver data. These controllers outperform a PID controller in simulations, showing promise for lateral control applications. Moreno [36] presents a model-free control strategy for lateral vehicle control across various speeds, achieving accurate trajectory tracking, system stability, and passenger comfort, as demonstrated in simulation and real vehicle tests. Lastly, a survey paper [37] delves into deep learning technologies in autonomous driving, offering insights into their applications, strengths, and limitations.

This paper's structure is as follows: Section 2 explains clothoids' role in smooth path generation and the implementation of a clothoid-based tracking controller. Section 3 details the importance and the mathematical approach for determining adaptive lookahead distance for navigation. Section 4 addresses the constraints on steering rate and their implications on vehicle dynamics, while Section 5 details the design and validation of the lead filter, a key contribution for mitigating inherent lateral dynamics lag. The results of the implemented controller, assessing its tracking performance and behavior in various driving scenarios are thoroughly examined in Section 6. Finally, Section 7 concludes the study, reflecting on its contributions and future research directions.

## 2. Overview of the Clothoid-Based Lateral Control

Clothoids are an effective solution for vehicle path generation due to their ability to maintain continuous curvature. The primary benefit of using clothoids is their capacity to create paths that are smooth in position, direction, and curvature. Our study employs a previously described method [3], which incorporates clothoid construction into the tracking controller's functionality. At each time step, the controller computes the optimal clothoid from the vehicle's current position to a predetermined lookahead point on the reference path, leading the vehicle along a hyper-clothoid trajectory. These generated clothoids facilitate the calculation of the steering wheel angle for the upcoming time step. Furthermore, this method is characterized by its low computational demands and ease of real-time application.

Figure 1 graphically depicts the controller's operation, showing vehicle positions over time and the corresponding clothoids generated by the controller. It is evident that as the cross-track error diminishes, the constructed clothoid increasingly aligns with the intended reference path.

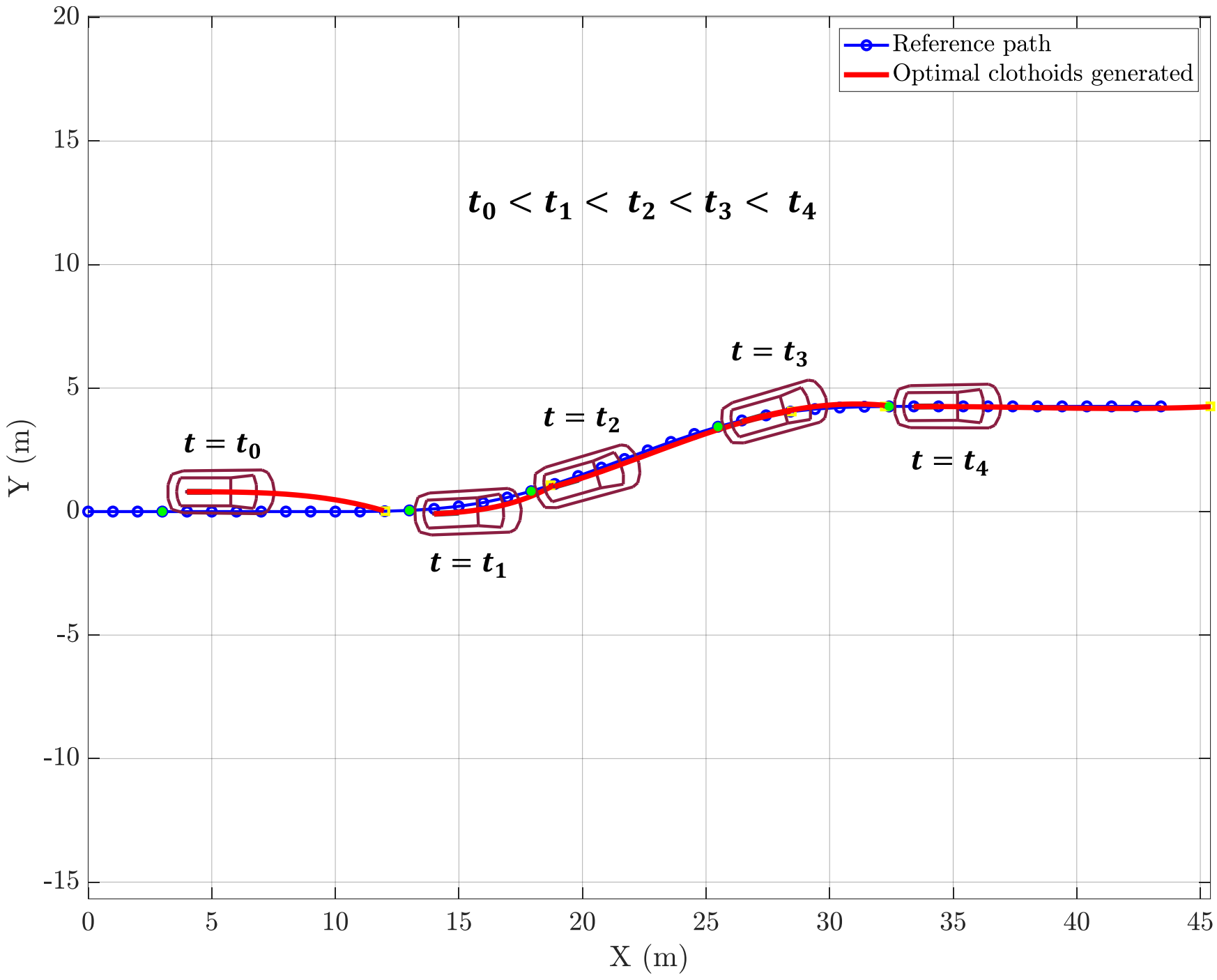

**Figure 1.** Sequential vehicle positions ($t_0$ to $t_4$) with the corresponding optimal clothoids generated by the lateral controller.

Regarding the foundational theory of clothoids, these curves are characterized by a curvature that varies linearly with respect to their arc length. Clothoids are formally expressed through parametric representations of Fresnel Integrals, specifically $S(t)$ for the sine component and $C(t)$ for the cosine component.

$$S(t) = \int_0^t \sin\left(\frac{\pi}{2}\tau^2\right)\, d\tau \tag{1}$$

$$C(t) = \int_0^t \cos\left(\frac{\pi}{2}\tau^2\right)\, d\tau \tag{2}$$

Theoretically, fitting clothoids between two points to create a smooth path depends on the specific requirements for the start and end conditions, such as orientation (direction) and curvature at those points, for example, for the following:

a. Two clothoids: Transitioning from one curve of a certain curvature to another curve of a different curvature, requires two clothoid segments—one to decrease the curvature to zero (straighten out) and another to increase the curvature to match the second curve.
b. Three clothoids: In more complex path-planning scenarios, such as an S-curve where the vehicle needs to transition from a curve in one direction to a curve in the opposite direction, a sequence of three clothoids might be used. The first clothoid segment adjusts the curvature from the initial curve to a straight path, the second segment transitions through zero curvature, and the third segment increases the curvature to match the final curve's requirements.

The presented approach for waypoint navigation employs hyper-clothoids, utilizing a single clothoid segment fit from the location of the car to a target point located on the reference path at a certain Euclidean distance in front of the vehicle. Within the context of path planning, the boundary value problem associated with this method is established with the primary objective of approximating the target point as closely as feasible, subject to the constraints imposed by the saturation limits of the curvature rate ($\kappa'$). Consequently, the boundary conditions are reformulated as detailed below:

$$\left.\begin{array}{ll} x(0) = x_{\text{car}}, & x(L) = x_{\text{target}} \pm \delta_x \\ y(0) = y_{\text{car}}, & y(L) = y_{\text{target}} \pm \delta_y \\ \psi(0) = \psi_{\text{car}}, & \kappa(0) = \kappa_{\text{car}} \end{array}\right\} \tag{3}$$

The clothoid segment from the location of the car is projected such that the path is drivable and is continuous with respect to the curvature and curvature rate. Due to these constraints on the clothoid projection, there may be instances where the segment does not terminate precisely at the target point. By introducing tolerances, denoted as $\delta_x$ and $\delta_{y,}$ it is acknowledged that achieving an exact endpoint match with the target point is not a prerequisite for the effectiveness of the proposed solution. This acknowledgment stems from the understanding that reaching the target point precisely in a single iteration is impractical. Therefore, the path from the current location to the target is formed using a single clothoid characterized by a constant curvature rate $\left(\kappa' = \frac{d\kappa}{ds}\right)$. For a comprehensive mathematical explanation and details on the generation of the implemented clothoid, readers are encouraged to consult reference [3].

In such waypoint navigation control systems, lookahead distance determines how a vehicle anticipates and tracks its desired path. Lookahead distance represents the distance ahead of the vehicle along its current trajectory at which specific target points or waypoints are placed. The selection of an appropriate lookahead distance is a critical decision as it influences the vehicle's behavior [38–40]. A major novelty of the current work over the previously devised method is the development of a more robust and mathematically sensible way of evaluating this lookahead distance for the most commonly used heuristic approach as discussed in [3,10].

This study employs a co-simulation framework, integrating TruckSim® (2017 edition) and Simulink® (Release 2022a) to create a simulated environment. Figure 2 illustrates the communication pathways for the closed-loop lateral control implemented within Simulink®. As indicated in the block diagram, the kinematic quantities exported from TruckSim® closely align with real-world data obtained through GPS, rather than abstract state variables. It is important to note that, in this implementation, the complete reference path coordinates or waypoints are available in advance. The following steps summarize the working of the implemented clothoid controller:

(1) Global waypoints (*NWPs* and *EWPs*) are established as the navigational path for the vehicle.
(2) The clothoid controller receives the vehicle's current kinematic state $(X, Y, V_x, V_y, \psi, \delta_{Steer})$ and the desired path (*NWPs*, *EWPs*).
(3) The clothoid controller computes the curvature rate ($\kappa'$) required to steer the vehicle towards the target path.
(4) The product block calculates the product of the curvature rate ($\kappa'$) and vehicle's speed ($V$) to determine the rate of change of the vehicle's curvature ($\mathrm{d}\kappa/\mathrm{d}s$).
(5) The Look-Up Table (LUT) or kinematic model utilizes the steering angle ($\delta_{Steer}$) to predict the vehicle's future state based on current steering dynamics.
(6) The lead filter compensates for the vehicle's steering dynamics lag, ensuring the steering input is timely and accurate, preventing oscillatory behaviors.
(7) The compensated steering angle is then fed into the TruckSim® vehicle model, which simulates the vehicle's response to the steering input.
(8) TruckSim® provides a real-time simulation environment, updating the vehicle's kinematic state $(X, Y, V_x, V_y, \psi, \delta_{Steer})$ based on the provided steering inputs and vehicle dynamics.
(9) The updated kinematic state of the vehicle is fed back into the clothoid controller, closing the loop for continuous real-time navigation control.

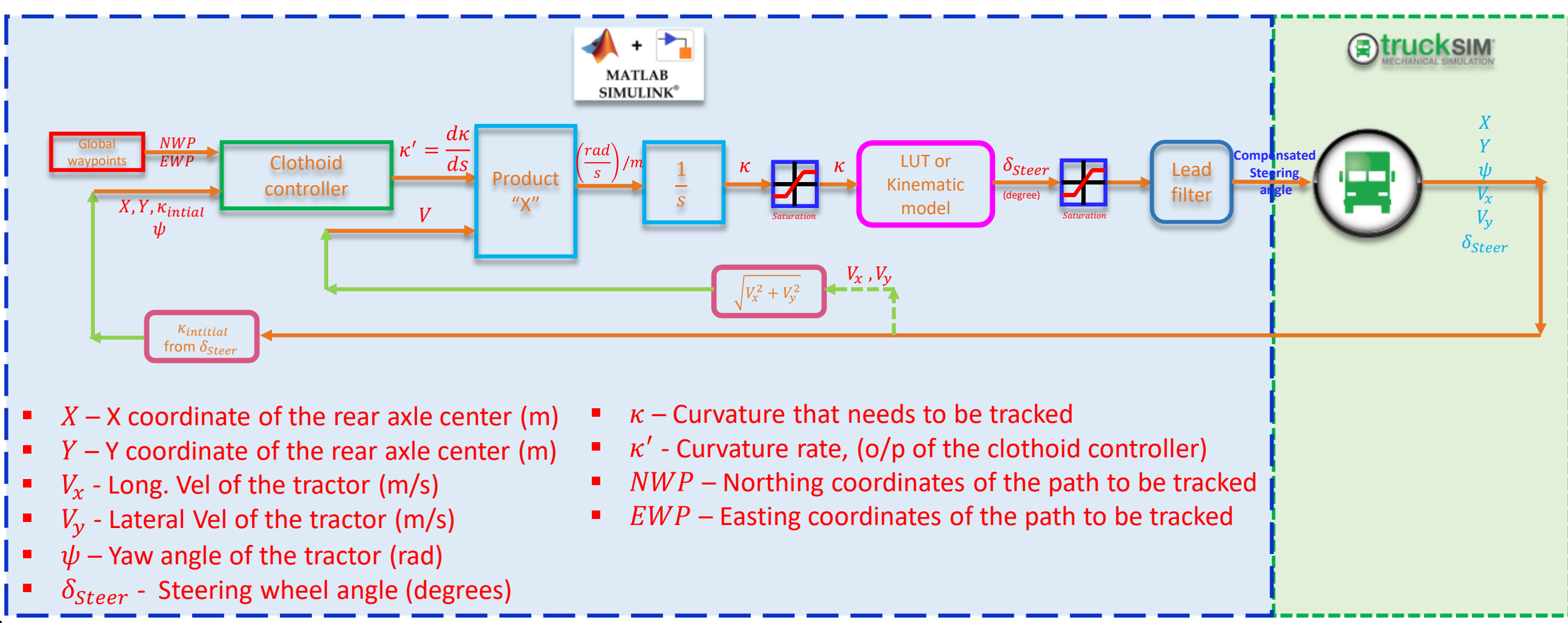

**Figure 2.** TruckSim®–Simulink® co-simulation framework for lateral control.

### 3. Adaptive Lookahead Distance

The concept of an adaptive lookahead distance holds significant importance in the domain of waypoint-based navigation for lateral control in autonomous vehicles. Unlike fixed lookahead distances, which rely on a predetermined value, an adaptive lookahead distance dynamically adjusts based on the vehicle's current situation and the geometry of the path to be tracked ahead. This adaptability is crucial because it allows autonomous vehicles to respond effectively to varying driving conditions, such as different road geometries, speeds, and traffic scenarios. By continuously evaluating and optimizing the lookahead distance, the vehicle can maintain a balance between path-tracking accuracy and stability. When navigating through complex scenarios, such as sharp turns or intersections, a shorter lookahead distance helps the vehicle make precise and timely steering adjustments. Conversely, on paths with small curvatures, a longer lookahead distance enables the vehicle to plan ahead and anticipate gradual lane changes.

In the past, researchers have used heuristic approaches [3,10], greedy algorithms [6], and even some fuzzy strategies [41] to estimate the lookahead distance at each time step. But, a more robust logical way of evaluating it would be to use mathematics to properly quantify the deviation between the two curves (in the case of clothoid control these two curves are (a) the clothoid curve fitted between the vehicle's current rear axle coordinates and the coordinates of the target point on the reference path to be tracked, which is at a distance from the vehicle characterized by the lookahead distance, and (b) part of the reference path between the closest waypoint to the vehicle's current position and the target waypoint). Then, we must choose the lookahead distance which gives a minimum deviation between the two curves. This is an iterative process, but the efficient implementation makes it possible for real-time use.

One of the mathematically valuable metrics employed in quantifying the dissimilarity between two curves or trajectories is "Fréchet distance". In the context of adaptive lookahead distance algorithms for waypoint navigation control, Fréchet distance plays a crucial role in assessing the disparity between the reference path and the vehicle's current trajectory. By measuring how closely the vehicle's projected path (clothoid between its current position and the target point) aligns with the reference path (part of the reference path between the closest waypoint to the vehicle's current position and the target waypoint) as shown in Figure 3, Fréchet distance aids in dynamically adjusting the lookahead distance. When the Fréchet distance is small, it indicates that the two curves are relatively close to each other, implying that the vehicle is effectively tracking the desired path. Conversely, a larger Fréchet distance suggests that the vehicle has deviated significantly from the reference path, prompting an adjustment in the lookahead distance to facilitate smoother and more precise path tracking.

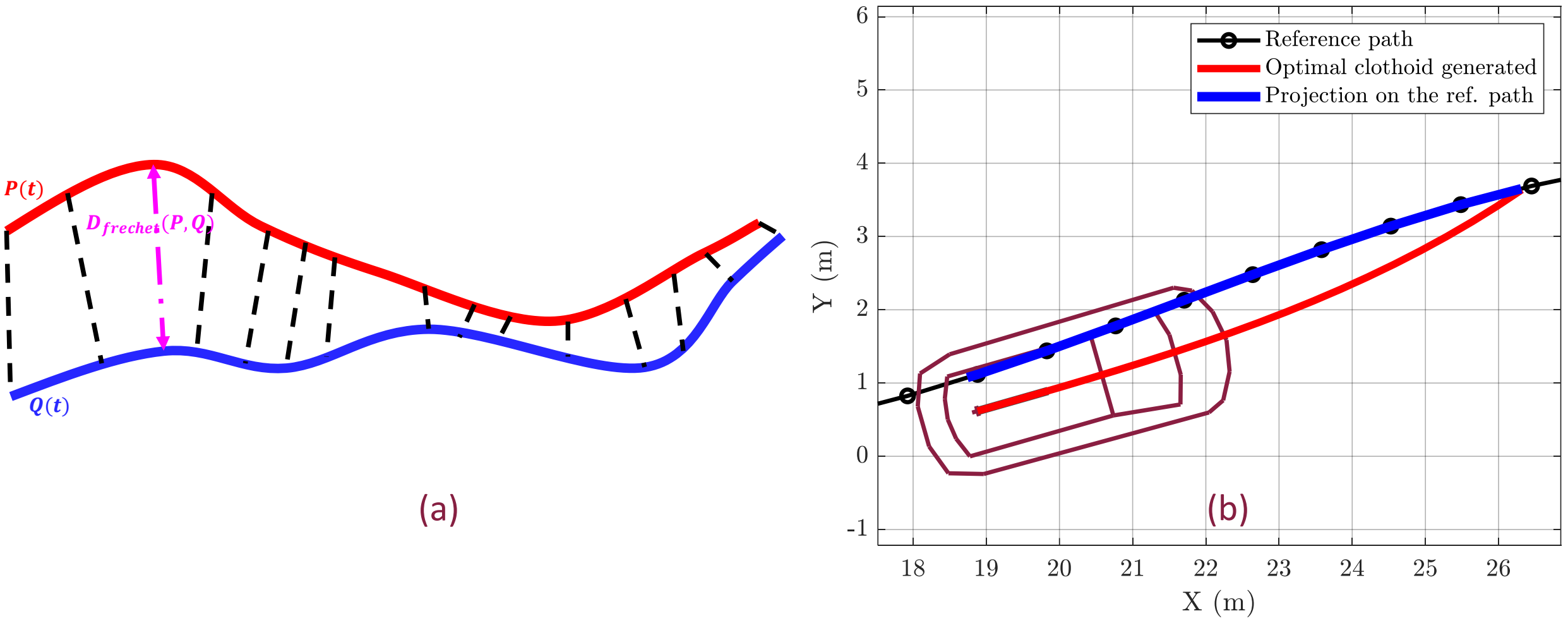

**Figure 3.** (**a**) Fréchet distance between curves $P(t)$ and $Q(t)$, representing the minimum leash length needed to connect points on the two paths (**b**) $P(t)$ and $Q(t)$ in the context of adaptive lookahead distance algorithm.

Mathematically, the Fréchet distance is defined as follows: given two parameterized curves, say $P$ and $Q$, both defined on a common interval (usually representing time or distance), the Fréchet distance measures the minimum "closeness" between these curves while allowing both curves to be traversed continuously. More precisely, it can be defined as the infimum (the greatest lower bound) of the maximum separation distance between corresponding points on $P$ and $Q$ as we vary over all possible continuous traversals of both curves. The key concept here is that both curves are considered to be traversed simultaneously along their respective parameterizations, and the Fréchet distance quantifies the minimum separation that can be achieved during such a traversal.

This concept is illustrated using Figure 3a and a simplified equation (Equation (4)) given below:

$$\mathrm{FD}(P,Q) = \inf_{\gamma,\delta} \max_{t}\{\mathrm{d}(P(\gamma(t)), Q(\delta(t)))\} \tag{4}$$

In Figure 3a, we have two curves, $P(t)$ and $Q(t)$, represented by the blue and red lines, respectively. These curves represent the paths of two objects (e.g., vehicles) over time. It is to be noted that calculating the exact Fréchet distance can be computationally intensive, and various algorithms and heuristics are used to approximate it efficiently in practice. Calculating the Fréchet distance in real-time implementation involves finding a discrete approximation of the continuous Fréchet distance between two trajectories. The implementation of the Fréchet distance evaluation in discrete time is provided in Appendix A.

Figures 4 and 5 underscore the significance of the method employed for dynamically updating the lookahead distance at each time step. This updating process relies on two key factors: the curvature of the reference path ahead, which the vehicle is expected to follow, and the vehicle's current positional deviation from this path (includes current cross-track error, heading error, etc.). In Figure 4b, it is evident that the Fréchet distance increases notably with the expansion of the lookahead distance, especially when the path's curvature is significant. This phenomenon arises due to the substantial dissimilarity between the clothoid trajectory fitted between the vehicle's present coordinates and its target destination (Figure 4a) and the reference path that the vehicle is intended to track. The observation that the Fréchet distance exhibits monotonic growth as the discrepancy between the locally fitted clothoid and the reference path increases enables the application of a threshold-based approach to ascertain an appropriate lookahead distance. This threshold value ultimately becomes one of the tunable parameters in the clothoid-based lateral controller.

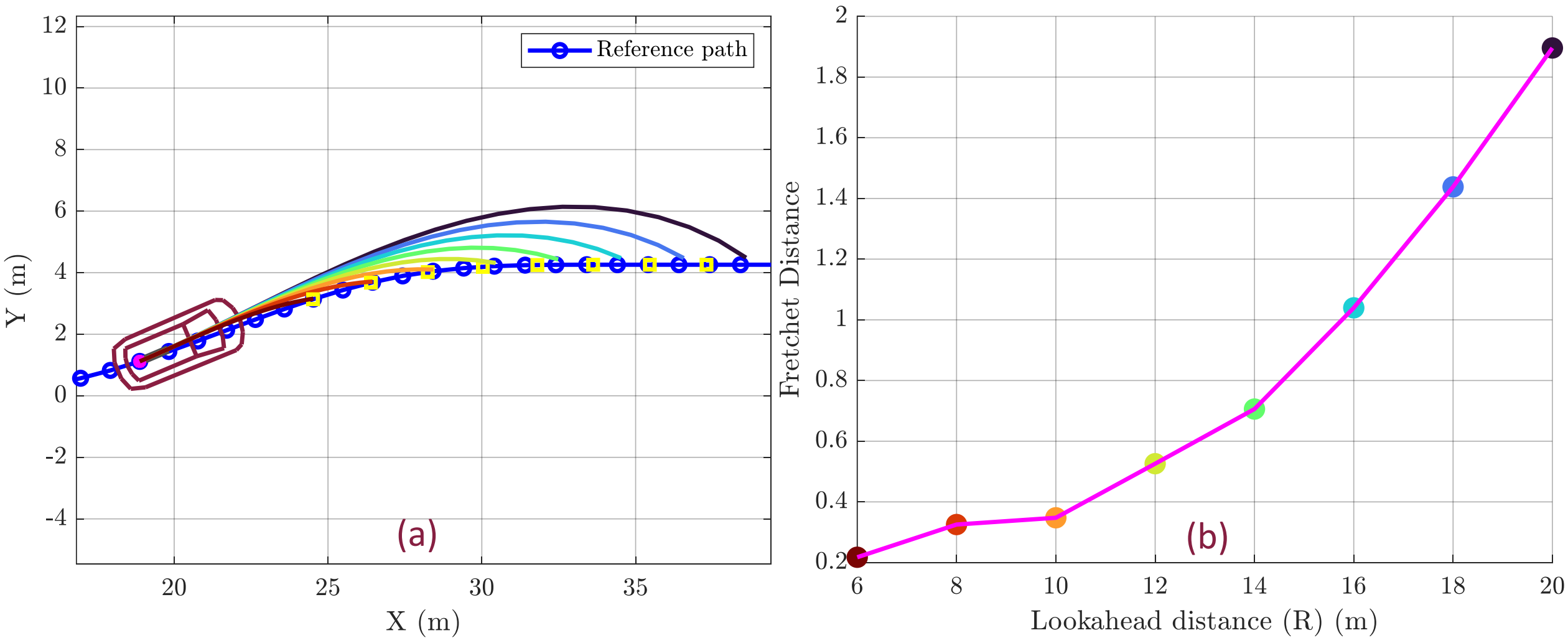

**Figure 4.** (**a**) Pictorial representation of deviation between the locally fitted clothoid and the reference path. (**b**) Variation of Fréchet distance with respect to the distance between the current vehicle coordinates and the target point.

Similarly, Figure 5 shows the variation in the minimum Fréchet distance (minimum of all the Fréchet distances obtained for lookahead distance varying from 5 to 20 m) with the initial cross-track error of the vehicle. Again, as the cross-track error increases, the deviation between the local clothoid fitted between the vehicle's current coordinates and the target point on the reference path and the part of the reference path between the closest point on the reference path, in relation to the vehicle's current position and the target point, increases. In addition to the initial cross-track error, the Fréchet distance will also depend upon the vehicle's current yaw angle and its initial curvature, since the clothoid controller uses that information too while obtaining the optimal clothoid curvature rate as illustrated in Figure 2.

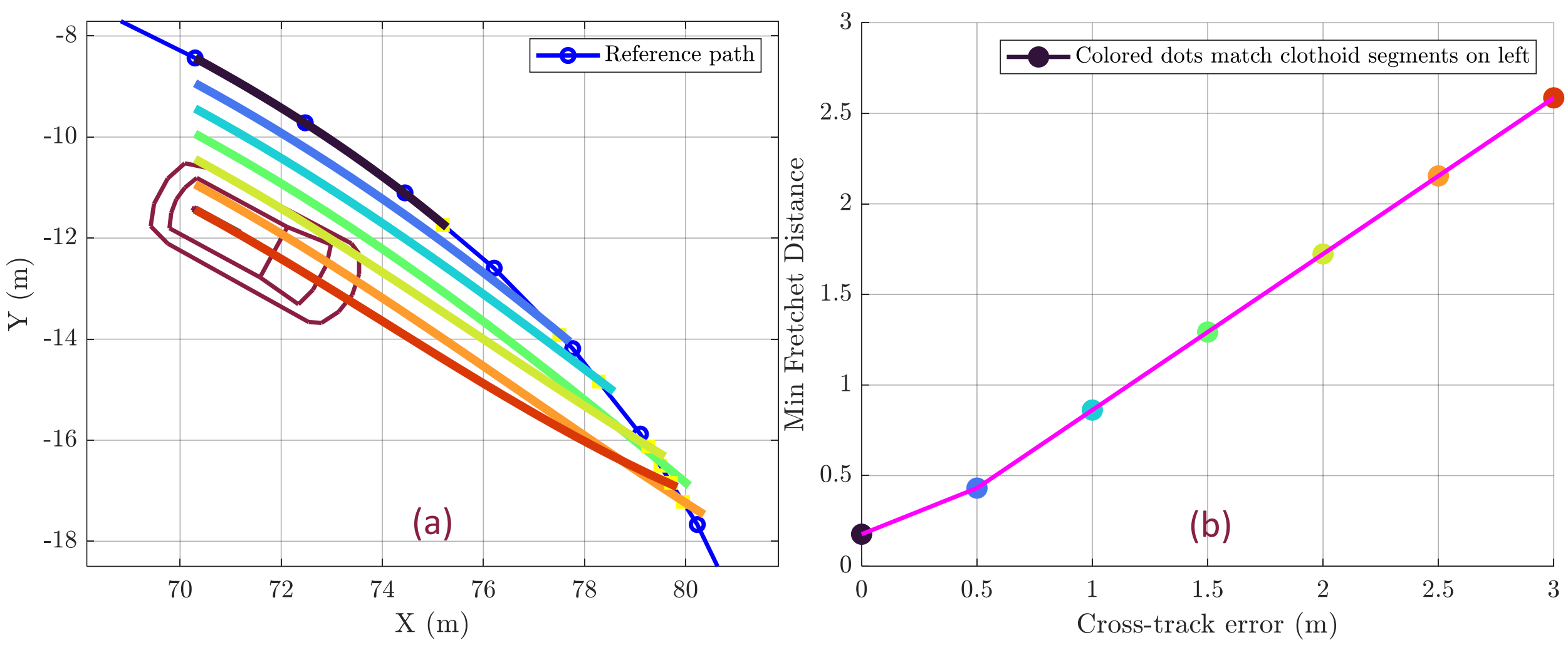

**Figure 5.** (**a**) Pictorial representation of deviation between the locally fitted clothoid and the reference path. (**b**) Minimum Fréchet distance vs. initial cross-track error.

After careful consideration of all these various factors (initial cross-track error, initial yaw angle, and initial curvature) and their influence on the Fréchet distance, we have determined a standardized threshold value of 0.5 m for the Fréchet distance through comprehensive simulation testing. This established threshold will be consistently applied in all subsequent lateral control simulations.

## 4. Curvature Rate Limits

Two factors contribute to the establishment of curvature rate limits. Firstly, vehicle dynamics imposes constraints on the maximum steering rate achievable at a given velocity. For heavily loaded vehicles, such as semi-tractor trailers, this limitation corresponds to the speed at which significant lateral G-forces can be induced, potentially leading to rollovers [42]. The second source of limitations arises from the specific clothoid controller algorithm employed in this study (for detailed information, see [3]). In the first scenario, curvature saturation has been incorporated into the integration block (refer to Figure 2), where it integrates the product of ($\kappa'$) (curvature rate) and the vehicle's speed $V$.

To discover these saturation limits, a sequence of step-steer tests was conducted at varying velocities, each involving distinct step-steering angle inputs. The saturation limits were established based on the curvature values corresponding to the steering angle input at which the vehicle experienced a rollover. The outcomes of this series of simulation tests, presented in Figure 6, confirm the expected trend of decreasing curvature limits as the vehicle's speed increases.

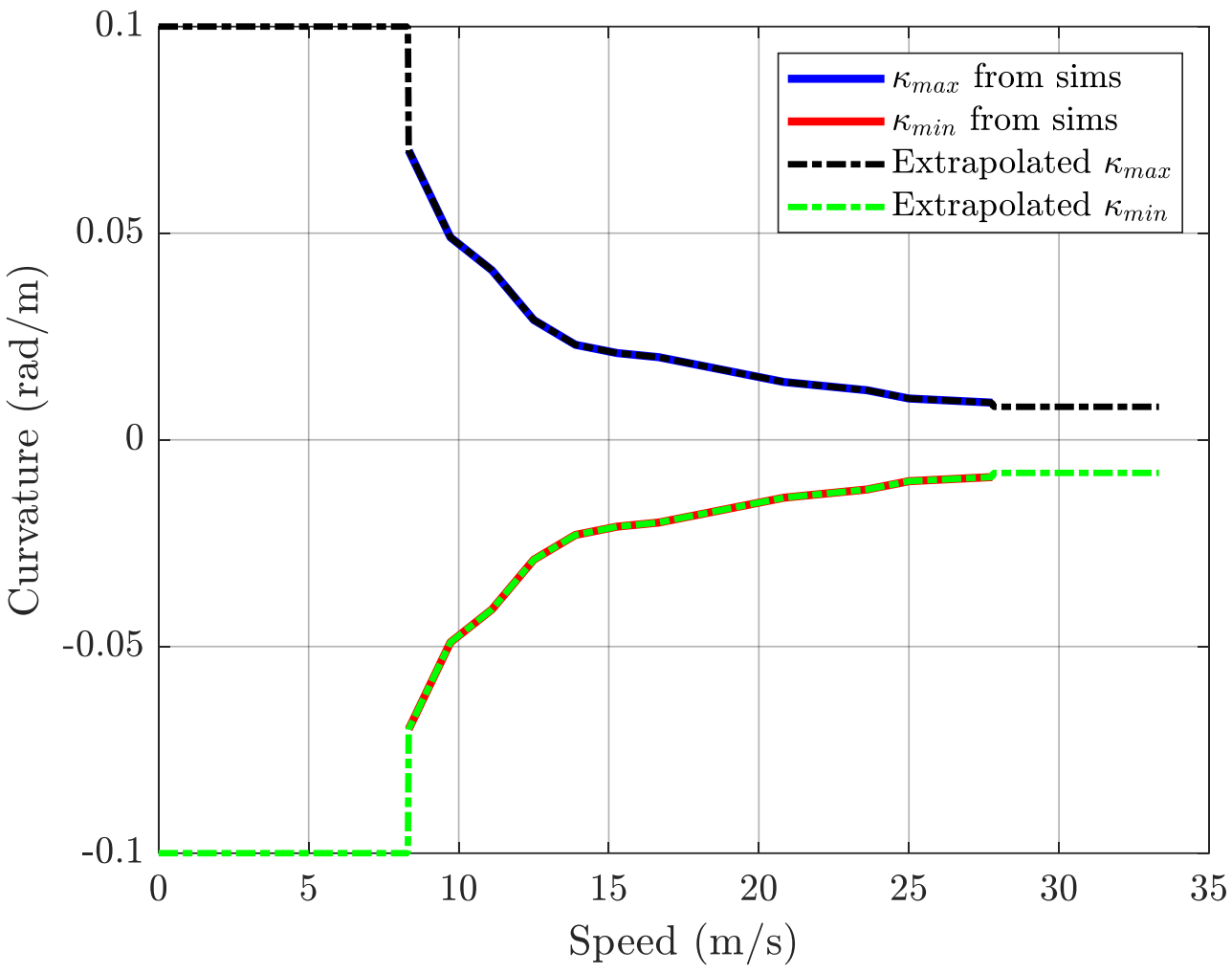

**Figure 6.** Curvature saturation limits based on roll-over speeds.

## 5. Design and Validation of Lead Filter

As mentioned in the previous section, there exists a lag between the application of the steering wheel angle (input) and the realization of a steady-state curvature of the vehicle's path traced in correspondance with the input. To compensate for this lag, a lead filter is designed, effectively increasing the bandwidth of the steering control system.

### *5.1. Estimation of the Lateral Dynamics Lag*

To design an effective lead filter, it is essential to initially characterize the original system's dynamics in terms of a transfer function. This characterization begins with a sensitivity analysis to identify the factors influencing the lag. Numerous simulations were conducted, maintaining a constant steering profile representative of a dual-lane-change (DLC) maneuver (a standardized test maneuver [43]). During these simulations, various parameters were modified, such as the payload mass and the desired velocity/speed for the entire maneuver. Figure 7 illustrates that, although the lag due to payload inertia is negligible, it is dependent on vehicle speed, with increased speed resulting in a greater lag.

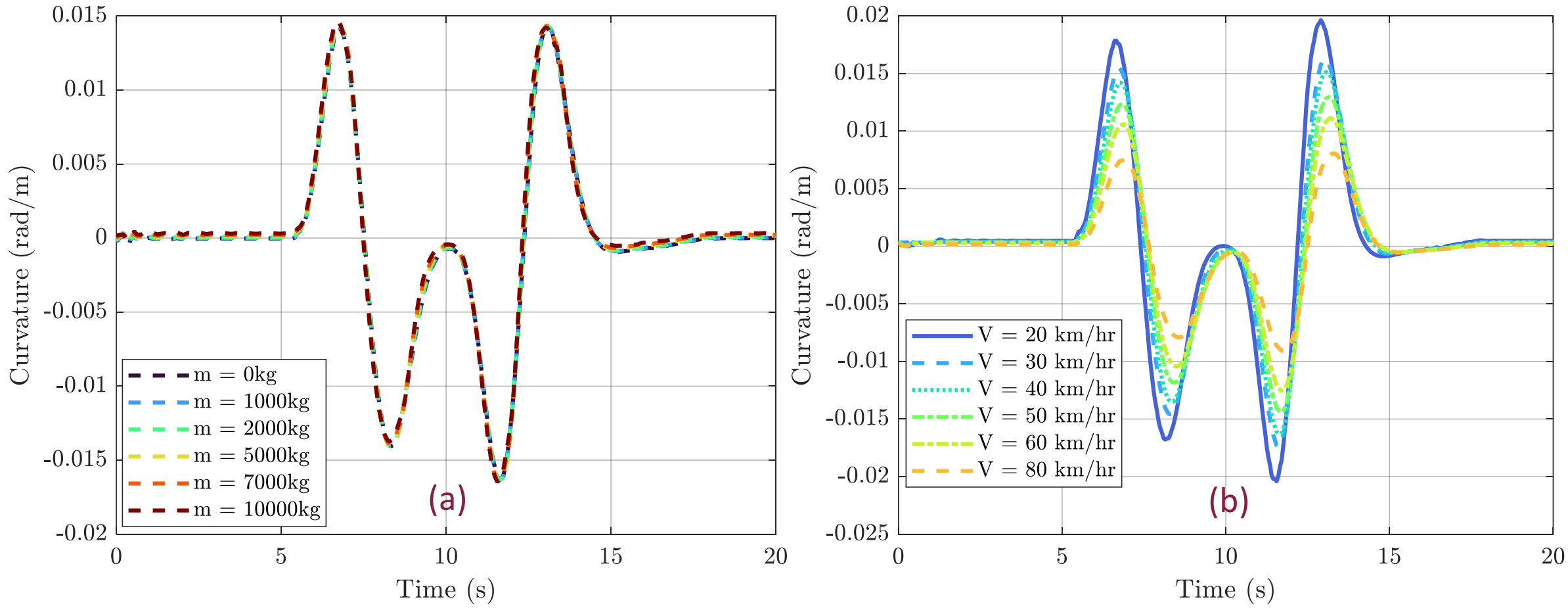

**Figure 7.** Variation of curvature w.r.t, (**a**) *payload mass*, (**b**) *vehicle speed* for the same steering profile input.

To characterize and reduce the lag, we use a second-order transfer function given using:

$$G_{lag}(s) = \frac{\omega_n^2}{s^2 + 2\zeta\omega_n s + \omega_n^2} \tag{5}$$

Here, $\omega_n$ represents the natural frequency and $\zeta$ represents the damping ration of the steering dynamics correspondingly. Both these parameters were fine-tuned using a comparative analysis with various simulation outcomes, as depicted in Figure 8.

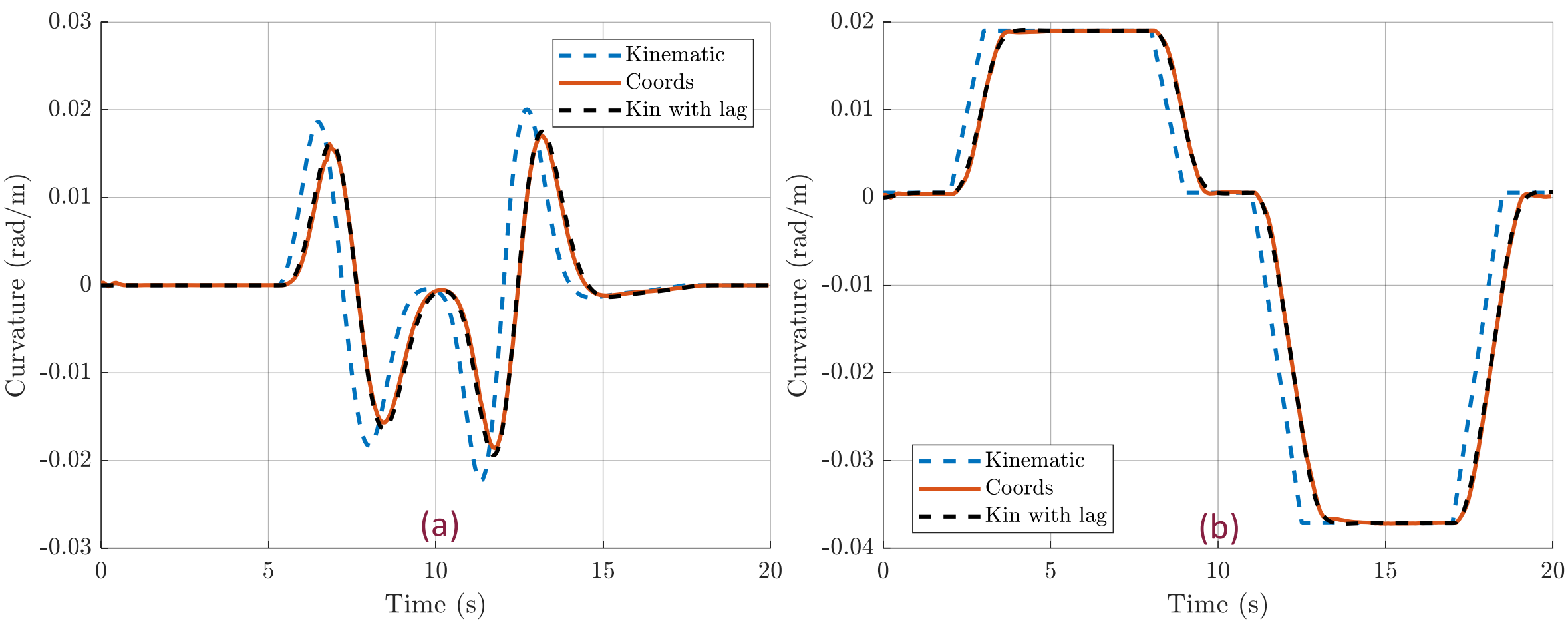

**Figure 8.** Characterization of a second-order-lag equation (transfer function) relating the steering wheel angle (i/p) to the realized curvature (o/p) when vehicle speed is ~30 kmph. (a) Dual-lane-change maneuver. (b) Series of step inputs.

Figure 8 illustrates the comparison among three curvature profiles: the curvature predicted using the simple bicycle model (derived from the Ackermann steering angle at the front, given using: $\mathbf{tan}\boldsymbol{\phi} = \boldsymbol{L}/\boldsymbol{R_1}$, where $\boldsymbol{\phi}$ −ackermann steer angle, $\boldsymbol{L}$ −wheelbase of the vehicle, and $\boldsymbol{R_1}$ −radius of curvature of the vehicle's rear axle center), the actual curvature computed from the rear axle coordinates of the vehicle, and the lagged curvature attained by filtering the kinematic model's output. As depicted, the proposed second-order transfer function effectively mirrors the inherent lag observed in TruckSim® simulation results, stemming from the vehicle's lateral/steer dynamics.

*5.2. Design and Validation of Lead Filter*

After characterizing the inherent lag in the lateral dynamics of the TruckSim® vehicle dynamic model, we used loop-shaping techniques to design a lead filter to compensate for the same. The problem was formulated with the transfer function representing the lag (Eqn. 5) as the plant and developing a compensator plant to increase the bandwidth of the combined system. The feed forward compensator, designed using loop-shaping techniques to increase the bandwidth of the cascaded system, consists of one pair of complex zero and three real poles as given below:

$$G_{lead} = \frac{\omega_p^3\,(s^2 + 2\psi\omega_z s + \omega_z^2)}{\omega_z^2 (s + \omega_p)^3} \tag{6}$$

Figure 9 illustrates the time–domain validation of the lead filter's lag compensation capabilities. Figure 9a,b depicts the compensated response for a dual lane change (DLC) as well as the step input steering maneuvers, respectively, as previously depicted in Figure 8. To delve into more detail, the **kinematic** relationship that links the steering wheel angle (input to TruckSim®) to curvature, essentially represents a normalized input. The curvature derived from the vehicle's coordinates signifies the actual curvature of the path followed by the vehicle's rear axle center. The '**Kinematic with lag**' plot portrays the outcome of the estimated lag (Equation (5)) coupled with the kinematic model (which maps the steering wheel angle to curvature, as mentioned above, $\tan\phi = L/R_1$). As demonstrated, this plot closely aligns with the actual curvature plot (derived from the path coordinates), which was its intended function.

The compensated response, represented by the pink-dotted-line plot, is obtained by initially filtering the steering wheel angle through the lead filter and subsequently passing it through the estimated kinematic model with lag. This plot effectively mitigates the lag and tends to align with the precise kinematic relationship output, which is the scaled-down input. This observation underscores the lead filter's role as a predictor, enhancing the signal to compensate for the lag introduced by TruckSim®'s lateral model. This compensation proves valuable when the lead filter is integrated with the clothoid lateral controller.

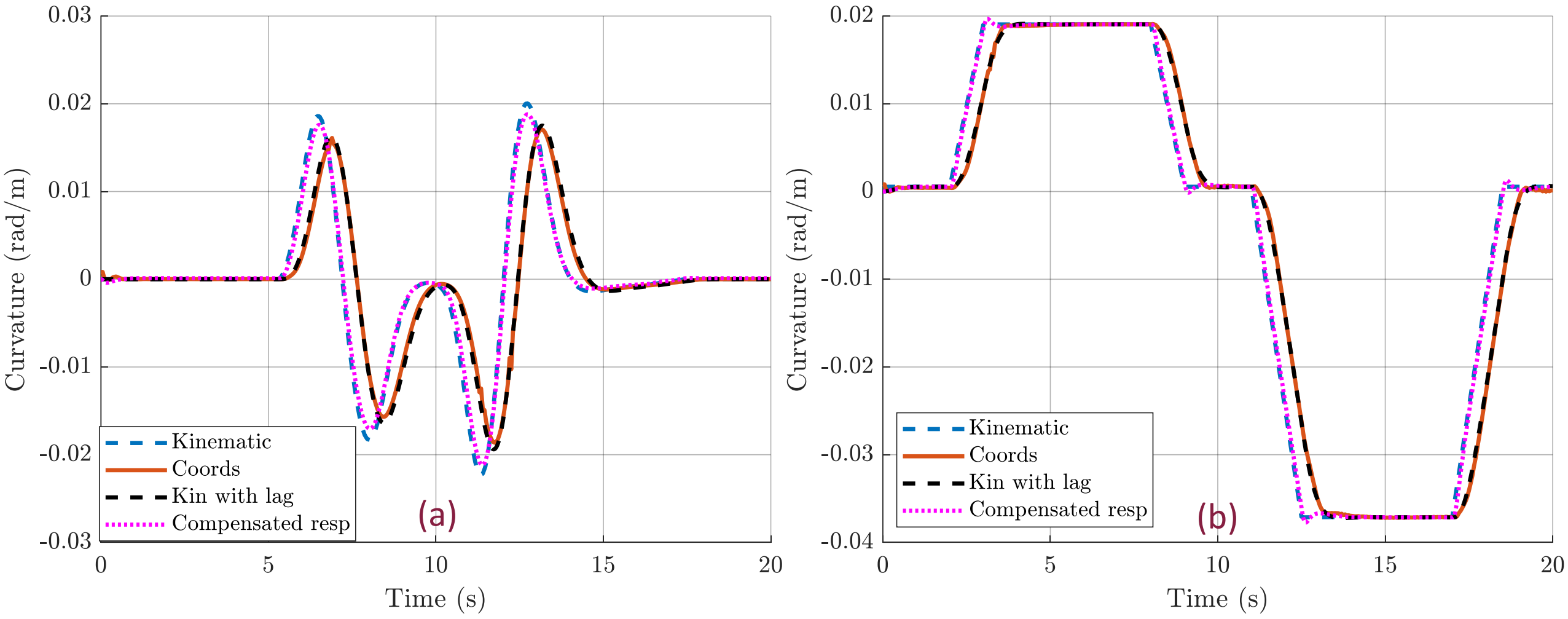

**Figure 9.** Time–domain validation of lead filter. (**a**) Dual-lane-change maneuver. (**b**) Step series maneuver.

Figure 10 shows the characteristics of the compensator (lead filter) dynamics, the original plant dynamics (estimated lag transfer function), as well as the cascaded dynamics. The increase in the total bandwidth of the system and the reduction in the phase lag in the desired frequency of interest can be inferred from Figure 10a and Figure 10b, respectively.

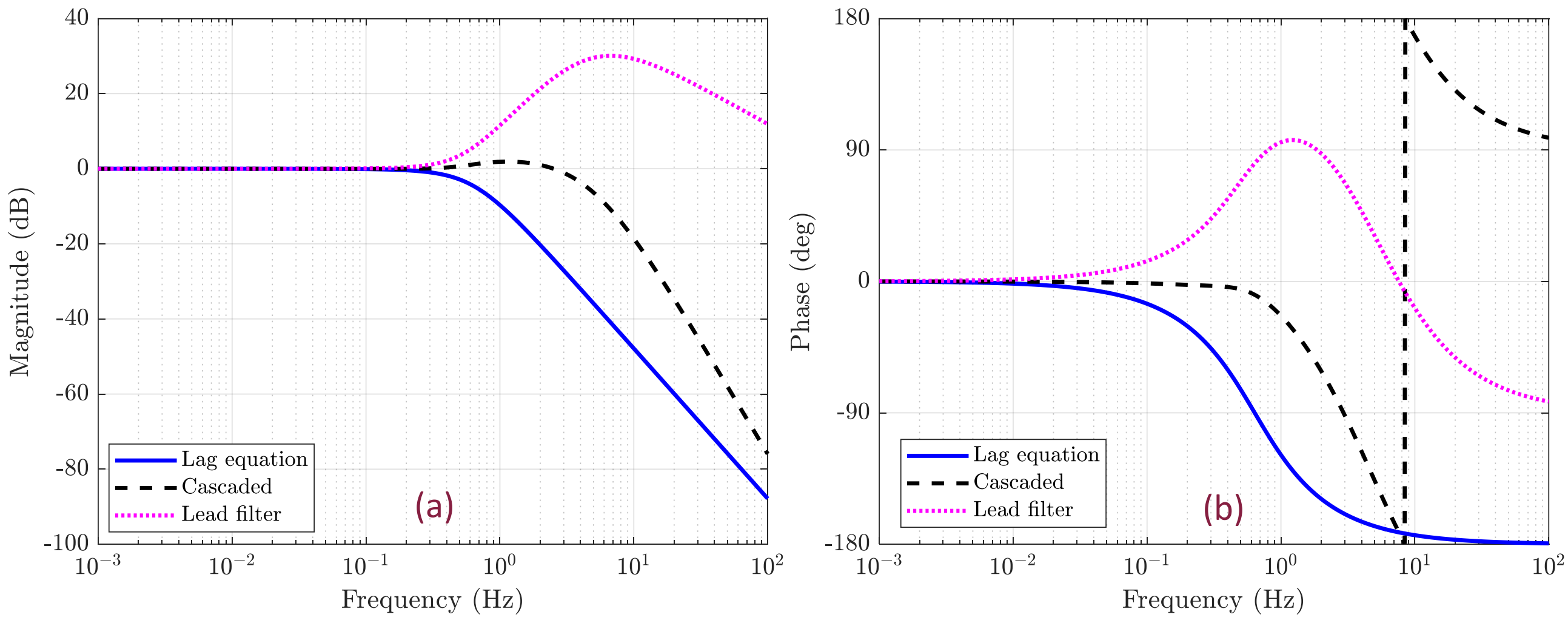

**Figure 10.** Frequency response of the plant (estimated lag transfer function), cascaded dynamics, and compensator (lead filter) (a) Magnitude and (b) Phase plot.

## 6. Results and Discussion

This section delves into the comprehensive results obtained from the implementation of a clothoid-based lateral controller featuring an adaptive Fréchet distance-driven lookahead adjustment mechanism. The objective here is to assess the controller's ability to maintain accurate lateral tracking across a diverse range of challenging curves. Additionally, the controller's effort and aggressive behavior will be scrutinized through steering wheel angle (control effort) versus time plots, shedding light on its effectiveness and stability under various simulated scenarios.

Notably, the lateral control simulations incorporated a speed controller to follow a reference velocity along the waypoints, crucial for realistic simulation in maintaining safe lateral acceleration, especially for a semi-tractor trailer. For this reason, the reference path dataset not only contains the northing and easting coordinates, but also the velocity/speed data at each of these waypoints (spatial velocity data). These velocity data were evaluated using "Smart velocity waypoint generation" which is a method for intelligently computing a batch of velocity waypoints based on given northing and easting waypoints. It emphasizes maintaining acceleration and velocity limits, ensuring the computed velocities are within a specified range and subject to constraints like lateral and longitudinal accelerations. This method uses quadratic polynomials to estimate curvature, heading, and path distance at each waypoint and adjusts velocities to optimize speed while adhering to the set constraints. The lateral acceleration constraint was set at 0.3 g's which is a conservative threshold for preventing rollover (refer [44,45]).

### 6.1. *Tracking Performance Assessment*

Precise lateral control is crucial for autonomous vehicles to adhere to their designated paths with minimal errors. The primary metric for evaluating the lateral controller's tracking performance is the cross-track error (CTE) plot. CTE quantifies the deviation of the ego vehicle (in this case, tractor's rear axle center) from their reference path and serves as a key indicator of how effectively the clothoid controller manages the lateral position.

The efficacy of the clothoid-based lateral controller is evident from tests conducted on various complex paths, including the dual-lane change and figure-eight maneuvers. The CTE plots for all the trajectories reveal low lateral deviations throughout the platooning maneuvers. This result underscores the effectiveness of the adaptive Fréchet distance-based lookahead distance mechanism in guiding the ego vehicle precisely along their reference paths.

### 6.2. *Controller Effort and Aggressiveness*

In addition to tracking accuracy, understanding the controller's effort and aggressiveness is crucial for assessing its stability and real-world applicability. One way to gauge these aspects is through the steering angle versus time plots, which provide insights into how smoothly the controller adjusts the steering inputs.

The steering angle versus time plots illustrate the dynamic response of the clothoid controller under various driving scenarios. It is essential to balance between aggressive maneuvers and maintaining the driver's comfort. Across all tested tracks, the controller consistently exhibited a well-behaved response, keeping the curvature changes within reasonable limits. This ensures that the steering inputs are smooth and gradual, minimizing abrupt changes that could potentially lead to discomfort for drivers and reduce the overall stability of the vehicle.

### 6.3. *Tracking of 8-Shaped Figure*

The figure-8 trajectory shown in Figure 11a combines tight turns with straightaways, requiring the vehicle to constantly adjust its steering angle and lateral acceleration. This tests its ability to handle a wide range of curvature demands. The continuous changes in curvature expose potential weaknesses in the control algorithms, revealing any instability or overcorrections in the steering system.

**Figure 11.** (**a**) Tracked trajectory of an 8-shaped figure. (**b**) CTE vs. time plot.

Figures 11 and 12 show the lateral control results obtained for an 8-shaped figure. Figure 11b represents the variation of cross-track error throughout the entire simulation run. The maximum cross-track error (CTE) observed throughout the entire simulation remained below 0.4 m, even during sharp turns. Examining the CTE data in more detail, it becomes apparent that the maximum CTE coincides with segments of the 8-shaped figure where the curvature rate $\left(\frac{d\kappa}{ds}\right)$ is at its peak. This correlation emphasizes the importance of considering local road geometry and curvature when determining the lookahead distance. The adaptive Fréchet distance-based mechanism effectively adjusts the lookahead distance, allowing the controller to anticipate and respond to varying curvature rates, thus minimizing the CTE.

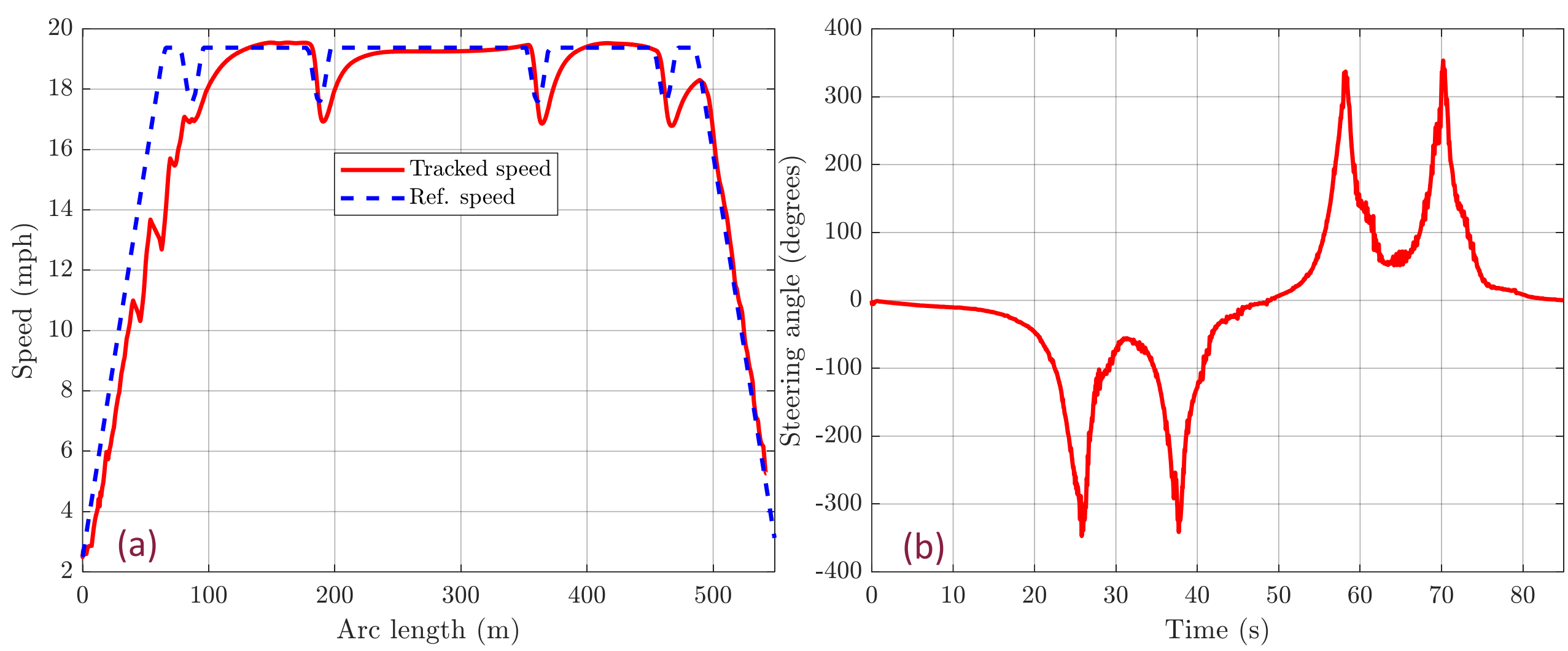

**Figure 12.** (**a**) Tracked speed vs. arc length. (**b**) Steering angle vs. time plot for the 8-shaped-figure tracking.

To understand the controller's behavior during the 8-shaped figure traversal, the steering wheel angle (control input to TruckSim®) versus time plot (Figure 12b) provides valuable insights. The lateral controller demonstrated a well-balanced response, keeping both curvature changes and resultant lateral acceleration within reasonable limits throughout the simulation. Figure 12a indicates the change in speed throughout the entire simulation. It can be noted that the vehicle negotiated the curves with a maximum curvature change ($d\kappa/ds$) at around 17 mph, a notable speed for such tight turning curves. This is further seen in the steering wheel angle vs. time plot Figure 12b which shows a maximum attained value of around 300 degrees at the points with maximum curvature changes, indicating the controller's ability to track highly demanding winding curves with a low cross-track error solidifying its tracking capabilities.

### 6.4. *Tracking of a Dual-Lane-Change (DLC) Maneuver Trajectory*

This subsection covers the lateral control results for the dual-lane-change maneuver trajectory. This maneuver mimics common driving scenarios encountered on highways and freeways, making the test results more applicable to real-world driving conditions. It encompasses various driving scenarios within a single test, including straight stretches, gradual lane changes, and adjustments while in the other lane. This diversity challenges the vehicle's lateral control system to adapt to different curvature demands. Figure 13a displays a near-perfect overlap of the tracked coordinates with the intended path during a dual-lane-change (DLC) maneuver, indicating a high degree of accuracy from the lateral controller. The accompanying CTE versus time plot in Figure 13b further demonstrates the controller's tracking capability, with cross-track error values fluctuating within a narrow range ($\pm$0.2 m), suggesting that the vehicle remained close to the desired trajectory throughout the maneuver. The minor deviations in CTE are typical of dynamic driving scenarios and reflect the controller's ability to correct the vehicle's path effectively over time.

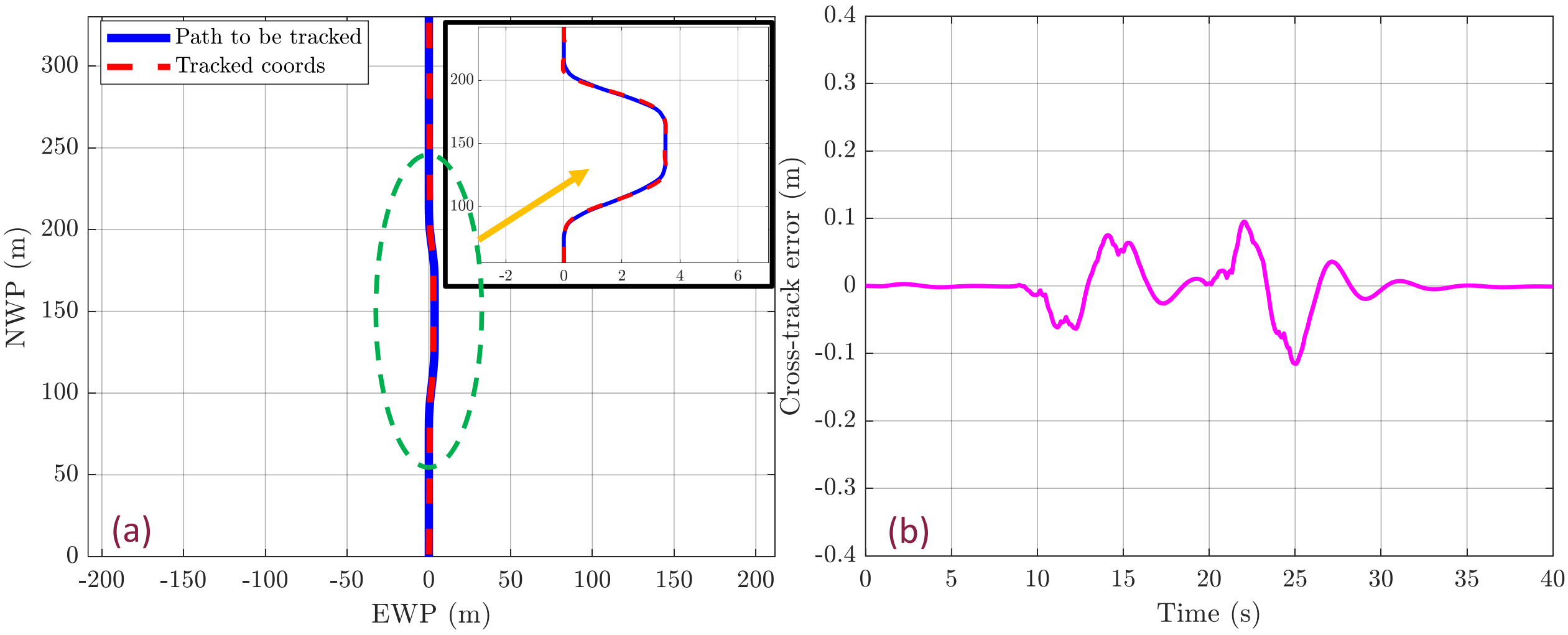

**Figure 13.** (**a**) Tracked trajectory of a dual-lane-change (DLC) maneuver. (**b**) CTE vs. time plot.

Figure 14b shows the steering angle over time, where the controller's output demonstrates a series of adjustments, reflecting active steering input to navigate the DLC trajectory. While the steering angles exhibit some peaks, suggesting responsive maneuvers to maintain the trajectory, the general trend indicates the controller's capability to output smooth steering commands appropriate for the speeds maintained, as shown in Figure 14a.

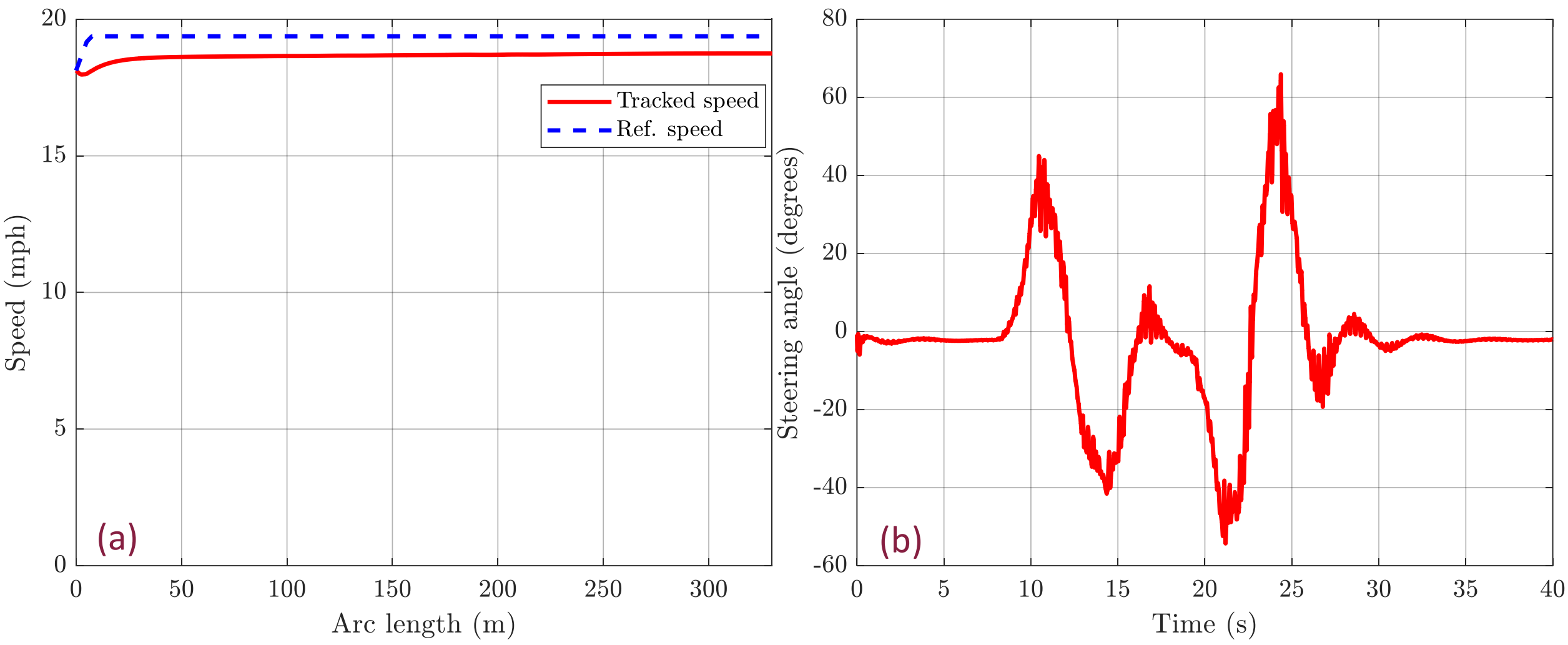

**Figure 14.** (**a**) Tracked speed vs. arc length. (**b**) Steering angle vs. time plot for the DLC trajectory tracking.

### 6.5. *Tracking of Zandvoort*

The Zandvoort track as shown in Figure 15a features a diverse range of corners, from tight hairpin turns to banked corners. This variety challenges the lateral control system of autonomous vehicles to adapt to different levels of curvature and adjust steering angles accordingly. This is crucial for testing the vehicle's ability to handle real-world driving scenarios with diverse cornering demands. In Figure 15 depicted below, the lateral controller's performance on the Zandvoort racetrack is showcased. Figure 15a illustrates the vehicle's tracked trajectory, closely mirroring the intended path, signifying the controller's precision. Figure 15b displays the cross-track error (CTE) over time, with the magnitude of deviations within a relatively narrow band $\pm 0.4$ m, indicating consistent tracking. The fluctuations in CTE suggest the controller's continuous adjustments in response to the racetrack's complex turns and straights, demonstrating its capability to handle both high-speed maneuvers and maintain the racing line with competence.

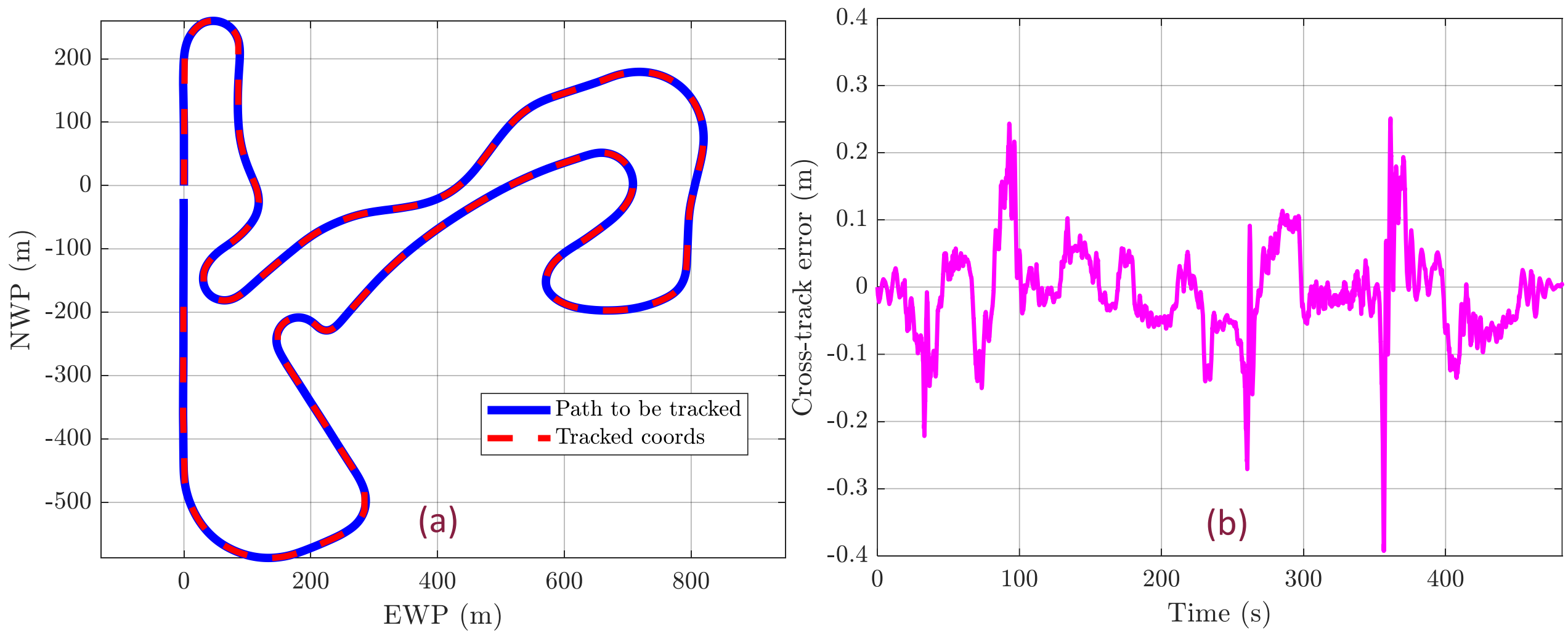

**Figure 15.** (**a**) Tracked trajectory of the Zandvoort racetrack. (**b**) CTE vs. time plot.

To gain insights into the controller's behavior during the Zandvoort racetrack simulation, we examine steering wheel angle versus time plots (Figure 16). These plots offer a dynamic view of the controller's responses to the track's twists and turns. The steering angle plot in Figure 16b exhibits a pattern of frequent adjustments, with some sharp spikes suggesting rapid steering corrections. These corrections correspond to the racetrack's complex turns, demonstrating the controller's active response to maintain the racing line. The degree of smoothness in the steering control output varies, with periods of gradually varying steering input interspersed with moments of aggressive steering to navigate the circuit.

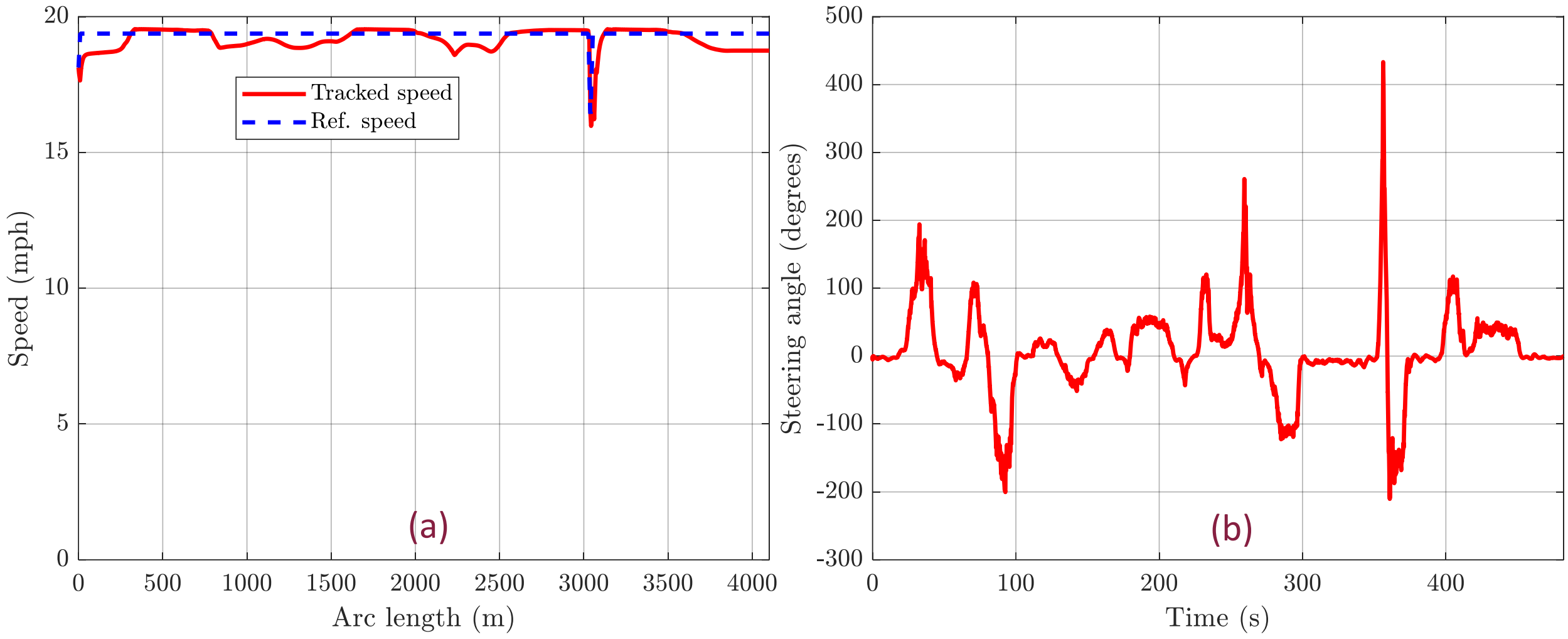

**Figure 16.** (**a**) Tracked speed vs. arc length. (**b**) Steering angle vs. time plot for the Zandvoort racetrack.

### *6.6. Comparative Analysis of Clothoid-Based Lateral Controller and Existing MPC-Based Lateral Controller*

This section presents a comparative analysis between the proposed clothoid-based lateral controller and an existing improved model predictive control (MPC)-based lateral controller [46].

The analysis reveals that the proposed clothoid-based lateral controller is closely competitive with the existing MPC-based lateral controller in terms of path-tracking accuracy as shown in Figure 17. It can be seen that the clothoid controller slightly delays in converging the error to zero when compared to the MPC approach. This observation may be ascribed to the differential testing conditions; the MPC controller was calibrated using a car model in CARSIM, which likely presents a less-complex steering trajectory than that required for a semi-tractor trailer. Nevertheless, the clothoid controller demonstrates a comparable level of tracking accuracy, as shown in Figure 17b.

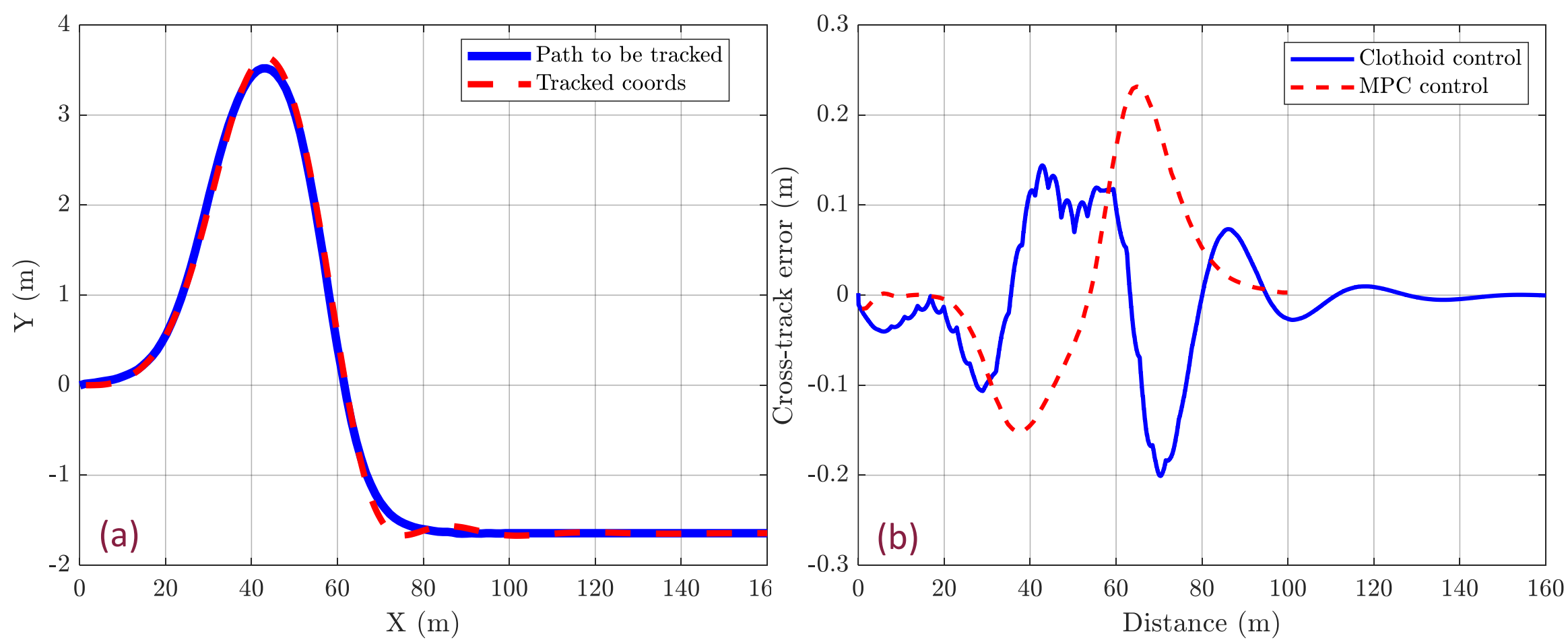

**Figure 17.** Comparative results with an existing MPC controller. (**a**) Tracked trajectory. (**b**) CTE vs. distance plot.

It is also to be noted that the MPC study did not specify the vehicle speed during trajectory tracking. To infer this, we extrapolated the approximate speed from their control effort (steering angle) versus time plot. The speed maintained by our clothoid-based controller, depicted in Figure 18a, is consistent despite the challenging curvature of the track. Further insights are gained from examining the steering angle versus time plot for the clothoid controller, as illustrated in Figure 18b. Although there are peaks in the steering angle, indicative of active responses required to maintain the designated path, the overarching pattern demonstrates the controller's proficiency in delivering smooth and measured steering inputs that align with the sustained velocities.

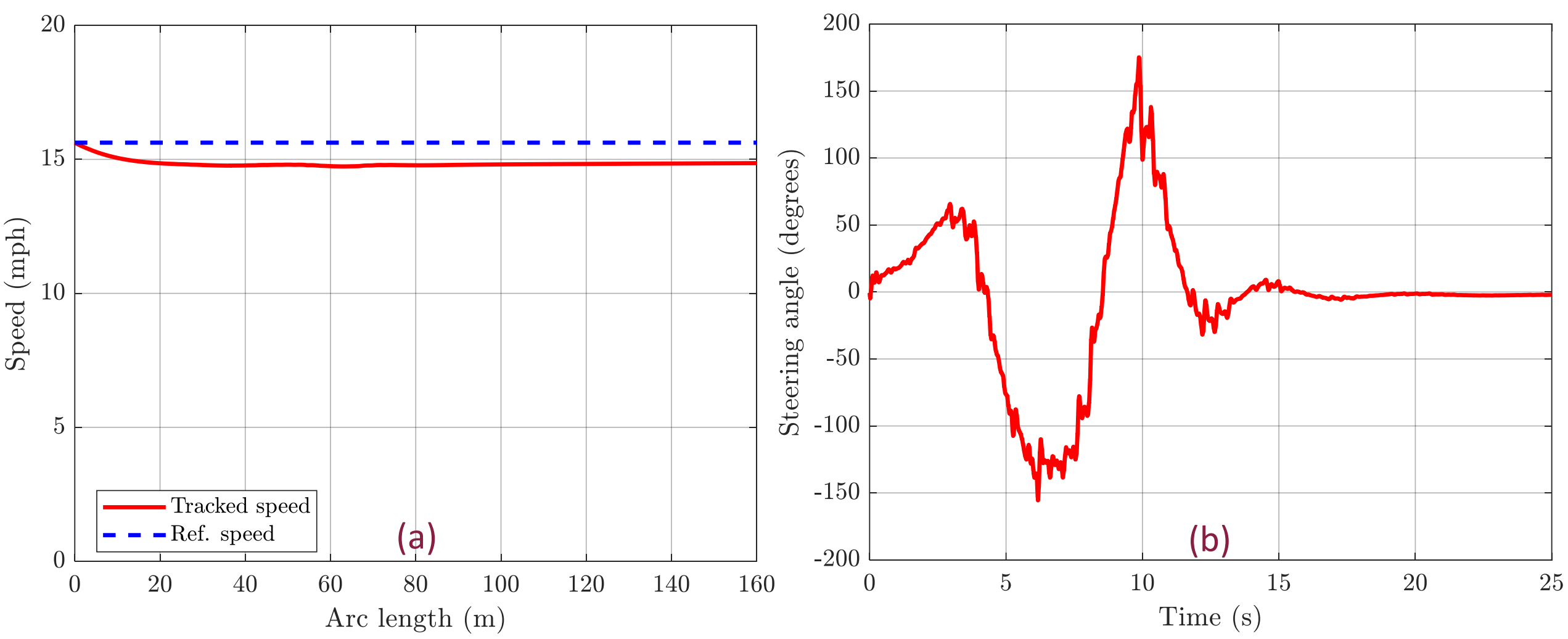

**Figure 18.** (**a**) Tracked speed vs. arc length. (**b**) Steering angle vs. time plot for the tracked trajectory.

## 7. Conclusions

The article discusses enhancing a clothoid-based lateral controller for autonomous vehicle navigation, focusing on path-planning and -control accuracy. The controller utilizes an adaptive Fréchet distance-driven lookahead strategy for precise trajectory tracking and maneuvers. The incorporation of a speed controller to complement the lateral control simulations added realistic constraints to the vehicle's dynamics, particularly in handling lateral accelerations preventing the semi-tractor trailer rollover. The findings indicate that the smart velocity waypoint generation method, in tandem with the clothoid-based approach, substantially improves path tracking. The results obtained from the implementation of the controller are promising. The controller consistently demonstrated good lateral tracking capabilities, at speeds of 30 kmph showcasing its ability to maintain precise control over challenging curves and maneuvers, such as figure-8 and dual-lane-change maneuvers, with minimal cross-track error. Additionally, the controller's well-behaved steering wheel angle output indicates its suitability for driver's comfort and overall vehicle stability, reinforcing its relevance for real-world applications. That said, the current configuration of the lead filter, tailored to the estimated lag transfer function for a specific speed (i.e., 30 km/h), presents suboptimal performance for high-speed tracking scenarios. Future work will extend the controller's capabilities to higher speeds and refine the closed-loop control to further minimize cross-track errors. The groundwork laid by this study paves the way for advanced autonomous vehicle systems capable of precise navigation and increased safety in complex driving scenarios.

**Author Contributions:** Conceptualization: A.S., S.S. and M.A.; methodology: A.S. and S.S.; software: A.S.; validation, A.S.; formal analysis: A.S., S.S., and M.A.; investigation: A.S. and S.S.; resources: M.A.; data curation, A.S.; writing—original draft: A.S.; preparation, A.S.; writing—review and editing, A.S., S.S. and M.A.; visualization, A.S. and S.S.; supervision: S.S. and M.A.; project

administration, M.A.; funding acquisition, M.A. All authors have read and agreed to the published version of the manuscript.

**Funding:** This research received no external funding.

**Institutional Review Board Statement:** Not applicable.

**Data Availability Statement:** The original contributions presented in the study are included in the article, further inquiries can be directed to the corresponding author.

**Conflicts of Interest:** The authors declare no conflicts of interest.

## Appendix A

Calculating the Fréchet distance in real-time implementation involves finding a discrete approximation of the continuous Fréchet distance between two trajectories. Here is a simplified explanation of how it can be found:

1. Discretize the Curves:
    1.1. First, you need to discretize the two curves, $P(t)$ and $Q(t)$, into a set of points. These points represent positions of the objects at specific time intervals or parameter values.
    1.2. For example, you can represent each curve as an array of points: $P = [(x1, y1), (x2, y2), \ldots]$ and $Q = [(u1, v1), (u2, v2), \ldots]$.
2. Dynamic Programming Matrix:
    2.1. Create a matrix, often referred to as the Fréchet distance matrix, to store intermediate results for calculations.
    2.2. The matrix is typically of size $(m + 1) \times (n + 1)$, where m is the number of points in curve $P$ and n is the number of points in curve $Q$.
    2.3. Initialize the matrix with a value that represents "infinity" for all entries except the top-left corner, which should be initialized to 0.
3. Fill the Matrix:
    3.1. Use dynamic programming to fill the matrix by calculating the Fréchet distance for all pairs of points in curves $P$ and $Q$.
    3.2. Start from the top-left corner and proceed row by row, column by column.
    3.3. For each cell (i, j) in the matrix, calculate the distance between points $P[i]$ and $Q[j]$.
    3.4. Update the cell $(i, j)$ with the maximum value of the following:
        - The distance between $P[i]$ and $Q[j]$.
        - The minimum distance in the adjacent cells to the left, above, and diagonally top left of the cell $(i, j)$.

    3.5. Continue this process until you reach the bottom-right corner of the matrix.
4. Final Result:
    4.1. The value in the bottom-right corner of the matrix represents the Fréchet distance between curves $P$ and $Q$.

The logic behind this approach is to find the minimum "closeness" between the two curves while allowing both to be traversed continuously. By considering all possible pairs of points between the two curves and using dynamic programming to efficiently compute the minimum distance, you can approximate the Fréchet distance.